# Multi-view Clustering with the Cooperation of Visible and Hidden Views

Ruixiu Liu, Zhaohong Deng, *Senior Member, IEEE*, Te Zhang,
Peng Xu, Kup-Sze Choi, Bin Qin, Shitong Wang

*Abstract*—Multi-view data are becoming common in real-world modeling tasks and many multi-view data clustering algorithms have thus been proposed. The existing algorithms usually focus on the cooperation of different views in the original space but neglect the influence of the hidden information among these different visible views, or they only consider the hidden information between the views. The algorithms are therefore not efficient since the available information is not fully excavated, particularly the otherness information in different views and the consistency information between them. In practice, the otherness and consistency information in multi-view data are both very useful for effective clustering analyses. In this study, a Multi-View clustering algorithm developed with the Cooperation of Visible and Hidden views, i.e., MV-Co-VH, is proposed. The MV-Co-VH algorithm first projects the multiple views from different visible spaces to the common hidden space by using the non-negative matrix factorization (NMF) strategy to obtain the common hidden view data. Collaborative learning is then implemented in the clustering procedure based on the visible views and the shared hidden view. The results of extensive experiments on UCI multi-view datasets and real-world image multi-view datasets show that the clustering performance of the proposed algorithm is competitive with or even better than that of the existing algorithms.

## I. INTRODUCTION

Clustering belongs to unsupervised learning paradigm. It is a process of dividing objects or data samples into different groups. As a kind of important data processing technique, clustering has a wide range of applications, including data mining, image processing and pattern recognition. Many clustering analysis methods have been proposed, such as K-means [1-3], Fuzzy C-means (FCM) [4-6], DBSCAN [7], OPTICS [8] and Spectral Clustering [9, 10] and so on. However, the data collected for clustering analyses are becoming more and more complex as the advancement of media technology. It poses great challenges to the traditional clustering methods. A typical issue is that datasets can be described by different attributes sets, i.e., multi-view datasets. Multi-view datasets are indeed common in real-world applications. For example, a document can be translated into two languages; a web page can be represented with two views, one as texts in a web page and the other as hyperlinks that point to the web page from other web pages. Traditional clustering methods mainly deal with single view data. To deal with multi-view data, traditional clustering methods usually consider each view independently, and then apply a simple ensemble mechanism [11, 12] to obtain the final results. Thus, traditional clustering methods usually ignore the connections between different views and fail to make full use of the information of the multi-view datasets, thereby exhibiting poor clustering performance for multi-view data. To meet with this challenge, multi-view learning technology is required. Different from traditional single view clustering algorithms, multi-view clustering methods integrate the information from different views to achieve the better performance. Hence, multi-view learning has become an important topic in the field of machine learning.

Researchers have proposed many algorithms to solve multi-view clustering problems. Based on the classical K-means algorithm, a two-level variable automatic weighted clustering algorithm called TW-k-means was proposed [13]. In order to solve the problem of large-scale multi-view data clustering, a multi-view K-means clustering method was proposed [14] which is robust to the outliers and can learn the weight of each view adaptively.

Based on the classical FCM algorithm, a large number of multi-view clustering algorithms have been proposed. By introducing collaborative mechanism into classical FCM, a collaborative clustering algorithm called Co-FC algorithm was developed in [15]. Based on FCM algorithm, a multi-view fuzzy clustering algorithm called Co-FKM was proposed in [16], which reduced the disagreement between the partitions on different views by introducing a penalty term in the objective function. A multi-view fuzzy clustering algorithm called Co-FCM was also proposed based on the classical FCM algorithm in [17]. It was further developed into the multi-view weighted collaborative fuzzy clustering algorithm (WV-Co-FCM) by applying different weights to different views. A multi-view fuzzy clustering algorithm called MinimaxFCM was proposed by introducing the minimum-maximum optimization strategy into the classical FCM algorithm [18].

In order to cope with the difficulties in high dimensional data

This work was supported in part by the Outstanding Youth Fund of Jiangsu Province (BK20140001), National Natural Science Foundation of China (61772239, 61272210) and Hong Kong Research Grant Council (PolyU152040/16E) (Corresponding author: Zhaohong Deng).

R. Liu, Z. Deng, T. Zhang, P. Xu, B. Qin and S. Wang are with the School of Digital Media, Jiangnan University, Wuxi, China. (6161611004@vip.jiangnan.edu.cn, dengzhaohong@jiangnan.edu.cn, 511865360@qq.com, 6171610015@stu.jiangnan.edu.cn, qinbin_sd@126.com, wxwangs.t@aliyun.com)

K. Z. Choi is with the Center of Smart Health, School of Nursing, the Hong Kong Polytechnic University, Hong Kong, (e-mail: thomasks.choi@polyu.edu.hk).

clustering, a correlational spectral clustering algorithm based on kernel canonical correlation analysis was proposed in [19]. The algorithm first projects the multi-view data from different feature spaces to a common low-dimensional subspace. K-means or other clustering algorithm was then used to cluster the data in the low- dimensional space. Another multi-view clustering algorithm based on canonical correlation analysis was proposed in [20], where the algorithm first used canonical correlation analysis to project the multi-view data to a common low-dimensional subspace, and then used K-means or other clustering algorithms to cluster the generated low-dimensional data.

Some researchers apply non-negative matrix factorization technology to multi-view data clustering, and propose some multi-view clustering algorithms. A multi-view clustering algorithm based on joint non-negative matrix factorization was proposed in [21], where a joint non-negative matrix factorization method was used to normalize the coefficient matrix from each view into a common consistent matrix that was considered as a potential representation of the original data. K-means and other clustering algorithms were then used directly to cluster the consistent matrix. In [22], by applying matrix factorization technique to clustering results independently generated from various views, a simple and effective algorithm was proposed to combine data from multiple views. The effectiveness of the algorithm was proved by clustering multi-view text data. In [23], a partial multi-view clustering algorithm based on non-negative matrix factorization was proposed, which introduces cluster similarity and manifold retention constraints into a unified framework.

The aforementioned algorithms have provided some feasible solutions for multi-view clustering problems but many of them (e.g. [13-18]) only focus on the visible information provided by the visible views, and ignore the shared hidden information between the visible views. Meanwhile, other multi-view clustering algorithms [19-23] only focus on mining the shared hidden information of different views. In fact, both the shared hidden information and the individual information of different visible views can play an important role in the clustering process. How to effectively combine the shared information and individual information is a challenging topic in the research of multi-view clustering algorithms. To this end, a multi-view clustering algorithm developed by the cooperation of visible and hidden views is proposed in this study. The main contributions of this paper are as follows:

1) A method that extracts the shared hidden view from multi-view data is proposed by using non-negative matrix factorization.

2) A multi-view clustering algorithm integrating both the visible and hidden views is proposed.

3) The performance of the proposed algorithm is evaluated on both UCI multi-view datasets and real-world image multi-view datasets.

The rest of this paper is organized as follows. Section II briefly reviews K-means algorithm, non-negative matrix factorization and multi-view clustering. Section III first proposes a shared hidden view data extraction method for multi-view data and then a novel multi-view clustering algorithm with the cooperation of visible and hidden views is proposed. Section IV evaluates the performance of the proposed algorithm by conducting extensive experiments on multi-view datasets and comparing it with many existing clustering algorithms. Section V draws conclusions and gives the future work of the study.

## II. RELATED WORK

### A. K-Means

K-means [1-3] is a classical clustering algorithm. Attributed to its simplicity and strongly adaptive ability, K-means has been applied in various fields.

Given a dataset of $N$ samples, the corresponding matrix can be expressed as $\boldsymbol{X} = [\boldsymbol{x}_1, \boldsymbol{x}_2, ..., \boldsymbol{x}_N]$, where $\boldsymbol{x}_j \in R^d, j = 1, 2, ..., N$. Taking Euclidean distance as the similarity measure, data samples are clustered into $C$ ($2 \leq C \leq N$) clusters. The cluster centers are represented by matrix $\boldsymbol{Z} = [\boldsymbol{z}_1, \boldsymbol{z}_2, ..., \boldsymbol{z}_C]$ with $\boldsymbol{z}_i \in R^d, i = 1, 2, ..., C$ and the partitions are represented by matrix $\boldsymbol{U} = [u_{ij}] \in R^{C \times N}$, where the value of $u_{ij}$ is either zero or one, with $u_{ij} = 1$ indicating that the sample $j$ is clustered into cluster $i$.

The objective function of the classical K-means algorithm is defined as

$$P(\boldsymbol{U}, \boldsymbol{Z}) = \sum_{i=1}^{C} \sum_{j=1}^{N} u_{ij} \|\boldsymbol{x}_j - \boldsymbol{z}_i\|^2$$

$$s.t. \quad \sum_{i=1}^{C} u_{ij} = 1, u_{ij} \in \{0, 1\}, j = 1, 2, ..., N \quad (1)$$

By solving the above objective function, the iterative equations of the partition and the cluster center are given as follows.

$$\begin{cases} u_{ij} = 1, & if \ D_i \leq D_s \ for \ 1 \leq s \leq C \\ & where \ D_s = \|\boldsymbol{x}_j - \boldsymbol{z}_s\|^2 \\ u_{sj} = 0, & for \ s \neq i \end{cases} \quad (2)$$

$$\boldsymbol{z}_i = \frac{\sum_{j=1}^{N} u_{ij} \boldsymbol{x}_j}{\sum_{j=1}^{N} u_{ij}} \quad (3)$$

### B. Non-negative matrix factorization

Non-negative matrix factorization (NMF) [24, 25] is a dimension reduction technology that has been widely used in different fields in the past decades, e.g., pattern recognition and image engineering. The dimension of non-negative dataset can be reduced through NMF. Given a non-negative data matrix $\boldsymbol{X} = [\boldsymbol{x}_1, \boldsymbol{x}_2, ..., \boldsymbol{x}_N]$ with $m$ dimensional features and $N$ samples, NMF aims to obtain two non-negative matrix factors $\boldsymbol{W} \in R_+^{m \times r}$ and $\boldsymbol{H} \in R_+^{r \times N}$, so that $\boldsymbol{X} \approx \boldsymbol{WH}$, i.e., their product can approximately represent $\boldsymbol{X}$, where $r$ denotes the reduced dimension, $\boldsymbol{W}$ and $\boldsymbol{H}$ denote the basis matrix and coefficient matrix respectively. Therefore, NMF can be transformed to the optimization of the following problem,

$$\min_{\boldsymbol{W}, \boldsymbol{H}} \|\boldsymbol{X} - \boldsymbol{WH}\|_F^2$$

$$s.t. \quad \boldsymbol{W} \geq \boldsymbol{0}, \boldsymbol{H} \geq \boldsymbol{0} \quad (4)$$

where $\|\cdot\|_F$ represents the Frobenius norm.

Eq.(4) can be solved by the strategy in [26], and the update rules are as follows.

$$H_{i,j} = H_{i,j} \frac{\left(W^T X\right)_{i,j}}{\left(W^T W H\right)_{i,j}} \tag{5a}$$

$$W_{i,j} = W_{i,j} \frac{\left(X H^T\right)_{i,j}}{\left(W H H^T\right)_{i,j}} \tag{5b}$$

In recent years, NMF is commonly used in clustering analyses. For example, a document clustering method was proposed based on NMF for the terminology document matrix [26]. A graph regularization non-negative matrix factorization algorithm called GNMF was proposed in [27]. In GNMF, the geometric information of data was encoded by constructing an affinity graph to achieve the NMF with respect to the graph structure.

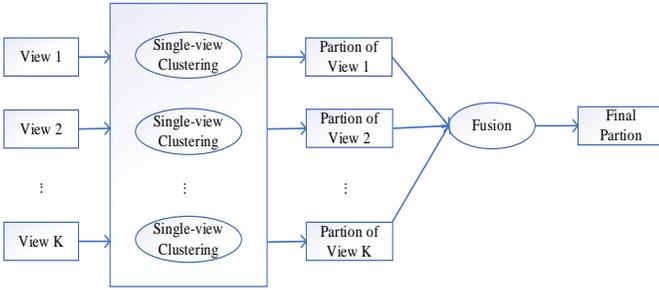

Fig. 1 A classical framework of single view clustering algorithms for processing multi-view data.

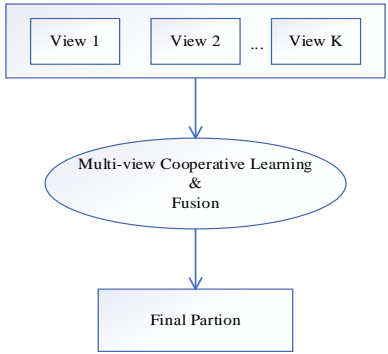

Fig. 2 A classical multi-view clustering framework.

## C. Multi-view Clustering

A typical framework for processing multi-view data using the traditional single view clustering algorithms is shown in Fig. 1. First, clustering is implemented independently to get the partition matrix of each view. The global partition matrix is then obtained by simple integration, such as weighting. Although this strategy provides a feasible way for processing multi-view data with single view clustering algorithms, the simple integration does not sufficiently consider the relevance between different views, which may result in poor clustering performance. Therefore, researchers attempt to improve multi-view clustering algorithms from different aspects. Fig. 2 shows a classical multi-view clustering framework. Multi-view learning aims to take full advantage of the information from different views and trains the model by considering the correlation and collaboration among the views. For example, the multi-view fuzzy clustering algorithms based on the FCM algorithm (Co-FKM) [16] and weighted views (WV-Co-FCM) [17] were thus proposed. Although the existing multi-view clustering algorithms can make full use of the relevance between different views, most of them only utilize the otherness information among different views, or only exploit the consistent information between the views. Therefore, the useful information of multi-view data is still not yet fully exploited for clustering analyses. How to effectively integrate otherness with consistency to implement multi-view clustering algorithms deserves in-depth study.

## III. MULTI-VIEW CLUSTERING WITH THE COOPERATION OF VISIBLE AND HIDDEN VIEWS

There are two key issues in multi-view learning: 1) how to make full use of the otherness information between different views of multi-view datasets, and 2) how to completely discover the consistency information between different views. From the last two sections, we can see that the existing multi-view learning methods usually only focus on the information of the visible views in the multi-view data, i.e., the otherness information is exploited more than the consistency information; or they only use the shared information for clustering, i.e., only the consistency information is used and the otherness information between different views is not considered.

In this section, a multi-view clustering algorithm with the cooperation of visible and hidden views, called MV-Co-VH, is proposed. The algorithm not only makes full use of the information of the visible views, but also the hidden information shared among different visible views. The framework of the MV-Co-VH is shown in Fig. 3. It can be seen that the algorithm not only achieves collaborative learning between visible views, but also collaborative learning between visible and hidden views. Therefore, the proposed MV-Co-VH algorithm utilizes the otherness and consistency information of the multi-view data simultaneously.

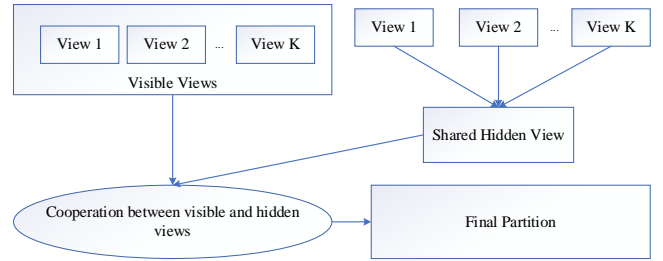

Fig. 3 The framework of MV-Co-VH.

### A. Shared Hidden View Extraction from Multi-view Data

For multi-view data, it is reasonable to assume that there is a hidden space shared by different visible views, and that the data of the different visible views can be generated from the shared hidden space data. This section presents how to extract the shared hidden view data using the NMF technique.

Given a multi-view dataset $\boldsymbol{X} = \{\boldsymbol{X}^1, \boldsymbol{X}^2, ..., \boldsymbol{X}^K\}$, the $k$th visible view is represented as the matrix $\boldsymbol{X}^k = [\boldsymbol{x}_1^k, \boldsymbol{x}_2^k, ..., \boldsymbol{x}_N^k] \in R^{m_k \times N}$, where $N$ is the number of samples, $K$ is the number of visible views, and $m_k$ is the number of features of the $k$th visible view. Since the elements in $\boldsymbol{X}^k$ may be negative, to use the NMF technique, we first normalize $\boldsymbol{X}^k$ so that the elements in $\boldsymbol{X}^k$ are non-negative. Let $\boldsymbol{H} = [\boldsymbol{h}_1, \boldsymbol{h}_2, ..., \boldsymbol{h}_N] \in R^{r \times N}$ be the data of multi-view dataset in the shared hidden space and $\boldsymbol{W}^k \in R^{m_k \times r}$ be a mapping matrix that maps the data of the hidden space to the $k$th visible view. The hidden view and mapping matrix can be obtained by solving the following optimization problems.

$$O = \sum_{k=1}^{K} q_k \|\boldsymbol{X}^k - \boldsymbol{W}^k \boldsymbol{H}\|_F^2 + \lambda \sum_{k=1}^{K} q_k \ln q_k$$

$$s.t. \begin{cases} \sum_{k=1}^{K} q_k = 1, 0 \leq q_k \leq 1 \\ \boldsymbol{W}^k, \boldsymbol{H} \geq \boldsymbol{0} \end{cases} \quad (6)$$

where the vector $\boldsymbol{q} = [q_1, q_2, ..., q_K]$ is a set of weights, $q_k$ is the weight of the $k$th visible view, and $\lambda \geq 0$ is a regularization coefficient. The first term of (6) indicates that the empirical loss of NMF. The second term corresponds to the adaptive weights of visible views based on the maximum entropy mechanism.

We can solve (6) by iterative optimization. The solution process can be divided into three steps: (i) optimizing $\boldsymbol{W}^k$ by fixing $\boldsymbol{H}$ and $\boldsymbol{q}$, (ii) optimizing $\boldsymbol{H}$ by fixing $\boldsymbol{W}^k$ and $\boldsymbol{q}$, and (iii) optimizing $\boldsymbol{q}$ by fixing $\boldsymbol{H}$ and $\boldsymbol{W}^k$. The details of the optimization are as follows.

(i) Optimizing $\boldsymbol{W}^k$ by fixing $\boldsymbol{H}$ and $\boldsymbol{q}$

When $\boldsymbol{H} = \hat{\boldsymbol{H}}$ and $\boldsymbol{q} = \hat{\boldsymbol{q}}$, the optimization problem becomes

$$\min_{\boldsymbol{W}^k} \sum_{k=1}^{K} q_k \|\boldsymbol{X}^k - \boldsymbol{W}^k \boldsymbol{H}\|_F^2 + \lambda \sum_{k=1}^{K} q_k \ln q_k \quad (7)$$

Therefore, $\boldsymbol{W}^k$ can be updated using the strategy in [25] and the update rule is as follows.

$$\left(\boldsymbol{W}^k\right)_{i,j} = \left[\frac{\left(\boldsymbol{X}^k \boldsymbol{H}^T\right)_{i,j}}{\left(\left(\boldsymbol{W}^k\right) \boldsymbol{H} \boldsymbol{H}^T\right)_{i,j}}\right] \left(\boldsymbol{W}^k\right)_{i,j} \quad (8)$$

(ii) Optimizing $\boldsymbol{H}$ by fixing $\boldsymbol{W}^k$ and $\boldsymbol{q}$

When $\boldsymbol{W}^k = \hat{\boldsymbol{W}}^k$ and $\boldsymbol{q} = \hat{\boldsymbol{q}}$, the optimization problem becomes

$$\min_{\boldsymbol{H}} \sum_{k=1}^{K} q_k \|\boldsymbol{X}^k - \boldsymbol{W}^k \boldsymbol{H}\|_F^2 + \lambda \sum_{k=1}^{K} q_k \ln q_k \quad (9)$$
$$s.t. \quad \boldsymbol{H} \geq \boldsymbol{0}$$

$\boldsymbol{H}$ can be updated using the strategy in [25] and the update rule is as follows.

$$\boldsymbol{H}_{i,j} = \left[\frac{\sum_{k=1}^{K} q_k \left(\left(\boldsymbol{W}^k\right)^T \boldsymbol{X}^k\right)_{i,j}}{\sum_{k=1}^{K} q_k \left(\left(\boldsymbol{W}^k\right)^T \boldsymbol{W}^k \boldsymbol{H}\right)_{i,j}}\right] \boldsymbol{H}_{i,j} \quad (10)$$

(iii) Optimizing $\boldsymbol{q}$ by fixing $\boldsymbol{H}$ and $\boldsymbol{W}^k$

When $\boldsymbol{H} = \hat{\boldsymbol{H}}$ and $\boldsymbol{W}^k = \hat{\boldsymbol{W}}^k$, the optimization problem becomes

$$\min_{\boldsymbol{q}} \sum_{k=1}^{K} q_k \|\boldsymbol{X}^k - \boldsymbol{W}^k \boldsymbol{H}\|_F^2 + \lambda \sum_{k=1}^{K} q_k \ln q_k \quad (11)$$

The update rule can be obtained based on the Lagrange multiplier method as follows.

$$q_k = \frac{\exp\left(-\frac{1}{\lambda} \|\boldsymbol{X}^k - \boldsymbol{W}^k \boldsymbol{H}\|_F^2\right)}{\sum_{h=1}^{K} \exp\left(-\frac{1}{\lambda} \|\boldsymbol{X}^h - \boldsymbol{W}^h \boldsymbol{H}\|_F^2\right)} \quad (12)$$

Based on the above analyses, the algorithm of extracting the shared hidden space data from the multi-view data by using NMF, called SHD-NMF, can be described below.

---

**Algorithm of SHD-NMF**

**Input:** A multi-view dataset $\boldsymbol{X} = \{\boldsymbol{X}^1, \boldsymbol{X}^2, ..., \boldsymbol{X}^K\}$ with $K$ visible views, the $k$th visible view $\boldsymbol{X}^k = [\boldsymbol{x}_1^k, \boldsymbol{x}_2^k, ..., \boldsymbol{x}_N^k] \in R^{m_k \times N}$, the desired reduced dimension $r$, parameter $\lambda$

**Output:** mapping matrices $\{\boldsymbol{W}^1, \boldsymbol{W}^2, ..., \boldsymbol{W}^K\}$ and the shared hidden view data $\boldsymbol{H}$

**Procedure**
1: Normalize each view $\boldsymbol{X}^k \ (1 \leq k \leq K)$
2: Initialize $\boldsymbol{W}^k$, $\boldsymbol{H}$, and $q_k \ (1 \leq k \leq K)$ randomly
3: repeat
4:   for $k = 1 : K$ do
5:     Update $\boldsymbol{W}^k$ by (8)
6:   end for
7:   Update $\boldsymbol{H}$ by (10)
8:   for $k = 1 : K$ do
9:     Update $q_k$ by (12)
10:  end for
11: until (6) converges

---

### B. Objective Function of MV-Co-VH

Given a multi-view dataset $\boldsymbol{X} = \{\boldsymbol{X}^1, \boldsymbol{X}^2, ..., \boldsymbol{X}^K\}$, the shared hidden view data of the visible views can be obtained by the SHD-NMF algorithm. Then, a multi-view clustering algorithm based on the cooperation of the visible and hidden views obtained, i.e., MV-Co-VH, can be developed. The objective function of MV-Co-VH is defined as follows.

$$J(\boldsymbol{U}, \boldsymbol{V}, \tilde{\boldsymbol{V}}, \boldsymbol{w}) = \beta \sum_{i=1}^{C} \sum_{j=1}^{N} u_{ij} \|\boldsymbol{h}_j - \tilde{\boldsymbol{v}}_i\|^2$$
$$+ (1-\beta) \sum_{k=1}^{K} w_k \sum_{i=1}^{C} \sum_{j=1}^{N} u_{ij} \|\boldsymbol{x}_j^k - \boldsymbol{v}_i^k\|^2 + \eta \sum_{k=1}^{K} w_k \ln w_k \quad (13)$$

$$s.t. \begin{cases} \sum_{i=1}^{C} u_{ij} = 1, u_{ij} \in \{0,1\}, 1 \leq j \leq N \\ \sum_{k=1}^{K} w_k = 1, 0 \leq w_k \leq 1 \\ \boldsymbol{H} \geq \boldsymbol{0} \end{cases}$$

where $\boldsymbol{U}$ is the partition matrix with the size of $C \times N$, whose elements $u_{ij}$ is a binary value with $u_{ij} = 1$ indicating that the sample $j$ is clustered in cluster $i$. $\boldsymbol{V} = \{\boldsymbol{V}^1, \boldsymbol{V}^2, ..., \boldsymbol{V}^K\}$ is a set of clustering centers with $K$ visible views, where $\boldsymbol{V}^k = [\boldsymbol{v}_1^k, \boldsymbol{v}_2^k, ..., \boldsymbol{v}_C^k]$ is the clustering center matrix of the $k$th visible view and $\boldsymbol{v}_i^k$ is the center of cluster $i$ of the $k$th visible view. $\tilde{\boldsymbol{V}} = [\tilde{\boldsymbol{v}}_1, \tilde{\boldsymbol{v}}_2, ..., \tilde{\boldsymbol{v}}_C]$ is the clustering center matrix of hidden view and $\tilde{\boldsymbol{v}}_i$ is the center of cluster $i$ of the hidden view. $\boldsymbol{w} = [w_1, w_2, ..., w_K]$ is a set of weights of visible views and $w_k$ is the weight assigned to the $k$th visible view. $\boldsymbol{H} = [\boldsymbol{h}_1, \boldsymbol{h}_2, ..., \boldsymbol{h}_N] \in R^{r \times N}$ is the shared hidden view of $K$ visible views. The terms in (13) are explained as follows.

(i) The first term $\sum_{i=1}^{C}\sum_{j=1}^{N} u_{ij} \|\boldsymbol{h}_j - \tilde{\boldsymbol{v}}_i\|^2$ and the second term $\sum_{k=1}^{K} w_k \sum_{i=1}^{C}\sum_{j=1}^{N} u_{ij} \|\boldsymbol{x}_j^k - \boldsymbol{v}_i^k\|^2$ are the sum of the within cluster dispersions of the hidden view and the sum of the within cluster dispersions of the visible views respectively. $0 \leq \beta \leq 1$ is a collaborative learning coefficient that is used to control the effectiveness of the two terms. Through collaborative learning of these two terms, the final partition matrix can be obtained.

(ii) In order to adaptively adjust the weight of each visible view, Eq. (13) introduces the Shannon entropy regularization term. Let $\sum_{k=1}^{K} w_k = 1$ and $w_k \geq 0$, by considering the weights of all visible views as a probability distribution, the Shannon entropy is expressed as $-\sum_{k=1}^{K} w_k \ln w_k$. The consistency of the weight of each view can then be obtained by minimizing the negative Shannon entropy $\sum_{k=1}^{K} w_k \ln w_k$. The regularization parameter $\eta$ is used to control the influence of the entropy.

## C. Optimization for Objective Function

We can minimize (13) by solving the following four subproblems iteratively:

Problem $P_1$: Set $\boldsymbol{V} = \hat{\boldsymbol{V}}$, $\tilde{\boldsymbol{V}} = \hat{\tilde{\boldsymbol{V}}}$ and $\boldsymbol{w} = \hat{\boldsymbol{w}}$, and solve the subproblem $J(\boldsymbol{U}, \hat{\boldsymbol{V}}, \hat{\tilde{\boldsymbol{V}}}, \hat{\boldsymbol{w}})$;

Problem $P_2$: Set $\boldsymbol{U} = \hat{\boldsymbol{U}}$, $\tilde{\boldsymbol{V}} = \hat{\tilde{\boldsymbol{V}}}$ and $\boldsymbol{w} = \hat{\boldsymbol{w}}$, and solve the subproblem $J(\hat{\boldsymbol{U}}, \boldsymbol{V}, \hat{\tilde{\boldsymbol{V}}}, \hat{\boldsymbol{w}})$;

Problem $P_3$: Set $\boldsymbol{U} = \hat{\boldsymbol{U}}$, $\boldsymbol{V} = \hat{\boldsymbol{V}}$ and $\boldsymbol{w} = \hat{\boldsymbol{w}}$, and solve the subproblem $J(\hat{\boldsymbol{U}}, \hat{\boldsymbol{V}}, \tilde{\boldsymbol{V}}, \hat{\boldsymbol{w}})$;

Problem $P_4$: Set $\boldsymbol{U} = \hat{\boldsymbol{U}}$, $\boldsymbol{V} = \hat{\boldsymbol{V}}$ and $\tilde{\boldsymbol{V}} = \hat{\tilde{\boldsymbol{V}}}$, and solve the subproblem $J(\hat{\boldsymbol{U}}, \hat{\boldsymbol{V}}, \hat{\tilde{\boldsymbol{V}}}, \boldsymbol{w})$.

(i) Problem $P_1$ can be solved as follows.

$$\begin{cases} u_{ij} = 1, & \text{if } D_i \leq D_s \text{ for } 1 \leq s \leq C \\ & \text{where } D_s = \beta \|\boldsymbol{h}_j - \tilde{\boldsymbol{v}}_s\|^2 + (1-\beta)\sum_{k=1}^{K} w_k \|\boldsymbol{x}_j^k - \boldsymbol{v}_s^k\|^2 \\ u_{sj} = 0, & \text{for } s \neq i \end{cases}$$

(14)

(ii) Problem $P_2$ and $P_3$ can be solved as follows.

$$\boldsymbol{v}_i^k = \sum_{j=1}^{N} u_{ij} \boldsymbol{x}_j^k \bigg/ \sum_{j=1}^{N} u_{ij} \tag{15}$$

$$\tilde{\boldsymbol{v}}_i = \sum_{j=1}^{N} u_{ij} \boldsymbol{h}_j \bigg/ \sum_{j=1}^{N} u_{ij} \tag{16}$$

(iii) The solution to problem $P_4$ is given by Theorem 1.

Theorem 1. Assuming that $\boldsymbol{U} = \hat{\boldsymbol{U}}$, $\boldsymbol{V} = \hat{\boldsymbol{V}}$ and $\tilde{\boldsymbol{V}} = \hat{\tilde{\boldsymbol{V}}}$, the necessary conditions for minimizing $J(\hat{\boldsymbol{U}}, \hat{\boldsymbol{V}}, \hat{\tilde{\boldsymbol{V}}}, \boldsymbol{w})$ is

$$w_k = \frac{\exp\left\{\dfrac{-(1-\beta)D_k}{\eta}\right\}}{\sum_{h=1}^{K}\exp\left\{\dfrac{-(1-\beta)D_h}{\eta}\right\}} \tag{17}$$

where $D_k = \sum_{i=1}^{C}\sum_{j=1}^{N} u_{ij} \|\boldsymbol{x}_j^k - \boldsymbol{v}_i^k\|^2$.

Proof. Refer to the objective function in (13), under the constraint $\sum_{k=1}^{K} w_k = 1$, the following Lagrange function can be established:

$$L(w_k, \gamma) = J(\hat{\boldsymbol{U}}, \hat{\boldsymbol{V}}, \hat{\tilde{\boldsymbol{V}}}, \boldsymbol{w}) + \gamma(\sum_{k=1}^{K} w_k - 1) \tag{18}$$

By taking the derivatives with respect to $w_k$ and $\gamma$ respectively, and setting the derivatives to be zero, the following equation can be obtained.

$$\frac{\partial L}{\partial w_k} = (1-\beta)\sum_{i=1}^{C}\sum_{j=1}^{N} u_{ij} \|\boldsymbol{x}_j^k - \boldsymbol{v}_i^k\|^2 + \eta(\ln w_k + 1) + \gamma = 0 \tag{19}$$

And

$$\frac{\partial L}{\partial \gamma} = \sum_{k=1}^{K} w_k - 1 = 0 \tag{20}$$

Furthermore, the following equation can also be obtained.

$$w_k = \frac{\exp\left\{\dfrac{-(1-\beta)}{\eta} \sum_{i=1}^{C}\sum_{j=1}^{N} u_{ij} \|\boldsymbol{x}_j^k - \boldsymbol{v}_i^k\|^2\right\}}{\sum_{h=1}^{K}\exp\left\{\dfrac{-(1-\beta)}{\eta} \sum_{i=1}^{C}\sum_{j=1}^{N} u_{ij} \|\boldsymbol{x}_j^h - \boldsymbol{v}_i^h\|^2\right\}} \tag{21}$$

## D. Algorithm Description

According to the rules of parameter learning deduced in the previous section, the pseudo-code of the proposed MV-Co-VH algorithm is given below.

---

**Algorithm of MV-Co-VH**

**Input:** A multi-view dataset $\boldsymbol{X} = \{\boldsymbol{X}^1, \boldsymbol{X}^2, ..., \boldsymbol{X}^K\}$ with $K$ visible views and the $k$th visible view given by $\boldsymbol{X}^k = [\boldsymbol{x}_1^k, \boldsymbol{x}_2^k, ..., \boldsymbol{x}_N^k] \in R^{m_k \times N}$, desired reduced dimension $r$, number of clusters $C(2 \leq C \leq N)$, convergence threshold $\varepsilon$, number of current iterations $t$, maximum number of iterations $T$, parameters $\beta$, $\eta$.

**Output:** Final partition matrix $\boldsymbol{U}$, cluster centers of different visible views $\boldsymbol{v}_i^k$, cluster centers of hidden view $\tilde{\boldsymbol{v}}_i$, weights of visible views $\boldsymbol{w} = [w_1, w_2, ..., w_K]$.

**Procedure**

**1:** Obtain the shared hidden view data $\boldsymbol{H} = [\boldsymbol{h}_1, \boldsymbol{h}_2, ..., \boldsymbol{h}_N] \in R^{r \times N}$ of the multi-view data by using the

SHD-NMF algorithm.

**2**: Randomly select the cluster centers of different visible views $\boldsymbol{v}_i^k (1 \leq i \leq C, 1 \leq k \leq K)$ and the cluster centers of hidden view $\tilde{\boldsymbol{v}}_i (1 \leq i \leq C)$, the random weights of visible views $\boldsymbol{w} = [w_1, w_2, ..., w_K]$ with $\sum_{k=1}^{K} w_k = 1$.

**3**: while $t \leq T$ do
**4**:   for $i = 1, 2, ..., C$
**5**:     for $j = 1, 2, ..., N$
**6**:       Update $u_{ij}$ by (14)
**7**:     end for
**8**:   end for
**9**:   for $k = 1, 2, ..., K$
**10**:     for $i = 1, 2, ..., C$
**11**:       Update cluster centers of different visible views $\boldsymbol{v}_i^k$ by (15)
**12**:     end for
**13**:   end for
**14**:   for $i = 1, 2, ..., C$
**15**:     Update cluster centers of hidden view $\tilde{\boldsymbol{v}}_i$ by (16)
**16**:   end for
**17**:   for $k = 1, 2, ..., K$
**18**:     Update weights of visible views $w_k$ by (17)
**19**:   end for
**20**:   if $\|J^{t+1} - J^t\| < \varepsilon$
**21**:     Algorithm stops;
**22**:   else
**23**:     $t = t + 1$
**24**:   end if
**25**: end while

## IV. EXPERIMENTAL STUDY

### A. Experimental Setup

The clustering performance of the proposed algorithm was evaluated experimentally using eight multi-view datasets, where five methods are from the UCI repository, one method is from the Caltech image multi-view dataset [26], one method is from the Corel image multi-view dataset [27] and one method is from Reuters dataset. The descriptions of these multi-view datasets are shown in Table II. In the experiments, the proposed MV-Co-VH algorithm was compared with seven selected clustering algorithms, i.e., fuzzy C-means clustering algorithm FCM [4], multi-view fuzzy clustering algorithm WV-Co-FCM [17], Co-FKM [16], MinimaxFCM [18], CombKM [28], MVKSC [29], and MultiNMF [21]. When single view clustering algorithms were concerned, single-view datasets were constructed by combining the features from all visible views directly.

The performance of the aforementioned clustering algorithms was evaluated using three performance indices, i.e., normalized mutual information (NMI) [2, 30], rand index (RI) [30, 31] and precision [32], which are defined as follow.

(i) Normalized Mutual Information (NMI)

$$NMI = \frac{\sum_{i=1}^{C}\sum_{j=1}^{C} n_{i,j} \log \frac{N \times n_{i,j}}{n_i \times n_j}}{\sqrt{(\sum_{i=1}^{C} n_i \log \frac{n_i}{N}) \times (\sum_{j=1}^{C} n_j \log \frac{n_j}{N})}} \quad (22)$$

(ii) Rand Index (RI)

$$RI = \frac{f_{00} + f_{11}}{N(N-1)/2} \quad (23)$$

(iii) Precision

$$Precision = \frac{f_{11}}{f_{00} + f_{11}} \quad (24)$$

Here, $n_{i,j}$ is the number of data points belonging to class $i$ and cluster $j$, $n_i$ is the number of data points in class $i$, $n_j$ is the number of data points in cluster $j$, $f_{00}$ denotes the number of any two data points belonging to two different clusters, $f_{11}$ denotes the number of any two data points belonging to the same cluster, and $N$ is the total number of samples in dataset. The values of these three indices are all within the interval [0, 1]. The higher the value is, the better the clustering performance is.

In the experiments, the appropriate settings of the parameters in the adopted algorithms were determined by grid search strategy. The range of search grid used is shown in Table I. With the optimal parameters determined, the algorithms were executed ten times, and the performance was reported in terms of the mean and standard deviation (SD) of NMI, RI and Precision.

### B. Experimental Results

As mentioned previously, eight multi-view datasets were used in the experiments. Five of them were obtained from the UCI repository, i.e., Multiple Features dataset, Image Segmentation dataset, Dermatology dataset, Forest Type dataset, and Water Treatment Plant dataset. The other three datasets were multi-view datasets Caltech, Corel and Reuters respectively. The Caltech dataset is an image dataset containing 101 classes, from which we selected three classes, i.e., Airplanes, Faces and Motorbikes, and preprocessed them into two-view data. The Corel dataset contains 34 classes of images, each class containing 100 pictures. We selected 10 classes with more prominent foreground and the images were represented with two views. Details of the multi-view datasets are shown in Table II. By using these seven datasets, the performance of the proposed MV-Co-VH algorithm was evaluated. The results of the experiments conducted on the eight multi-view datasets are shown in Tables III to X in terms of the mean and SD of NMI, RI and Precision. The results are also presented intuitively in terms of the mean of NMI, RI and Precision in Fig. 4 to Fig. 6.

The following conclusions can be drawn from the experimental results shown in Tables III to X and Figs. 4 to 6.

(i) The proposed MV-Co-VH algorithm shows the best performance in terms of NMI, RI and Precision on all the eight multi-view datasets.

(ii) The performance of the proposed MV-Co-VH algorithm is superior to that of the FCM and CombKM algorithms. For the two single view clustering algorithms, i.e., FCM, and CombKM, the results demonstrate that when the data samples are constructed by directly combining different views, the performance of these single view clustering algorithms are not satisfactory.

(iii) The performance of the proposed MV-Co-VH algorithm is better than that of the Co-FKM, WV-Co-FCM, MVKSC,

and MinimaxFCM on the multi-view datasets. While Co-FKM, WV-Co-FCM, MVKSC and MinimaxFCM only use visible views for clustering, MV-Co-VH further considers the hidden space information and the performance is improved. Compared with MultiNMF, MV-Co-VH algorithm achieves good clustering performance. MultiNMF algorithm only uses hidden information, but MV-Co-VH uses visible and hidden information at the same time.

To sum up, the proposed MV-Co-VH algorithm clearly demonstrates advantages over the single view algorithms on multi-view data. By considering of hidden views as well, the performance the proposed algorithm can be effectively improved and outperforms multi-view clustering algorithms that only utilize the visible views.

Table I Parameter setting of the algorithms under comparison.

| Algorithms | Parameters and setting |
|---|---|
| FCM | Fuzzy index $m$: $\{1.05,1.1,1.2,1.3,1.4,1.5,1.6,1.7,1.8,1.9,2.0\}$. |
| Co-FKM | Fuzzy index $m$: $\{1.05,1.1,1.2,1.3,1.4,1.5,1.6,1.7,1.8,1.9,2.0\}$; Collaborative learning parameter $\eta$: $[0,(K-1)/K]$, with step length 0.05 and the number of views $K$. |
| WV-Co-FCM | Fuzzy index $m$: $\{1.05,1.1,1.2,1.3,1.4,1.5,1.6,1.7,1.8,1.9,2.0\}$; Collaborative learning parameter $\eta$: $[0,(K-1)/K]$, with step length 0.05 and the number of views $K$. |
| MVKSC | Regularization parameter: $\{2^{-6},2^{-5},\cdots,2^{5},2^{6}\}$; Kernel parameter: $\{2^{-6},2^{-5},\cdots,2^{5},2^{6}\}$. |
| MinimaxFCM | Fuzzy index $m$: $\{1.05,1.1,1.2,1.3,1.4,1.5,1.6,1.7,1.8,1.9,2.0\}$; Parameter $\gamma$: $[0.1,0.9]$, with step length 0.1. |
| MultiNMF | Regularization parameter $\lambda_k$: $\{2^{-6},2^{-5},\cdots,2^{5},2^{6}\}$, $1\leq k\leq K$, where $K$ is the number of views. |
| MV-Co-VH | Regularization parameter $\eta$: $\{2^{-6},2^{-5},\cdots,2^{5},2^{6}\}$; Collaborative learning parameter $\beta$: $[0,1]$, with step length 0.1; Dimensionality of hidden view $r$: $\lceil 0.1*d\rceil,\lceil 0.2*d\rceil,\cdots,\lceil 1*d\rceil$, where $d$ is the minimum dimensionality of all visible views; if $d\leq 10$, the range of $r$ is $[1,\ d]$ with step length 1.0. |

Table II Description of eight multi-view datasets.

| Dataset | Size | Visible View 1 | | Visible View 2 | |
|---|---|---|---|---|---|
| | | Description | Number of Features | Description | Number of Features |
| Multiple Features | 2000 | *Fourier coefficients view*: Fourier coefficients of the character shapes | 76 | *Zernike moments view*: Zernike moments of the character shapes | 47 |
| Image Segmentation | 2310 | *Shape view*: shape information of the 7 images | 9 | *RGB view*: RGB values of the 7 images | 10 |
| Dermatology | 366 | *Histopathological view*: Histopathological information of 6 cases | 22 | *Clinical view*: Clinical information of 6 cases | 12 |
| Forest Type | 326 | *Image band view*: Image band information of the data | 9 | *Spectrum view*: Spectrum values and difference values | 18 |
| Water Treatment Plant | 527 | *Input view*: input conditions and input demands | 31 | *Output view*: output demands | 7 |
| Caltech | 2033 | *SIFT view*: SIFT features | 300 | *LBP view*: LBP features | 256 |
| Corel | 1000 | *SIFT view*: SIFT features | 300 | *LBP view*: LBP features | 256 |
| Reuters | 1200 | *Documents in English: English* | 2000 | *Documents in French: French* | 2000 |

Table III Performance of algorithms on the Multiple Features dataset (SD in bracket).

| Index | FCM | CombKM | Co-FKM | WV-Co-FCM | MVKSC | MinimaxFCM | MultiNMF | MV-Co-VH |
|---|---|---|---|---|---|---|---|---|
| NMI | 0.6807 (0.0146) | 0.6810 (0.0328) | 0.7020 (0.0147) | 0.6836 (0.0042) | 0.4323 (0.0206) | 0.6991 (0.0180) | 0.6729 (0.0118) | **0.7369 (0.0091)** |
| RI | 0.9201 (0.0065) | 0.9219 (0.0085) | 0.9332 (0.0030) | 0.9224 (0.0035) | 0.8770 (0.0068) | 0.9270 (0.0071) | 0.9256 (0.0044) | **0.9387 (0.0024)** |
| Precision | 0.6108 (0.0347) | 0.6012 (0.0444) | 0.6557 (0.0102) | 0.6116 (0.0166) | 0.3873 (0.0324) | 0.6256 (0.0393) | 0.6230 (0.0223) | **0.6822 (0.0169)** |

Table IV Performance of algorithms on the Image Segmentation dataset (SD in bracket).

| **Index** | FCM | CombKM | Co-FKM | WV-Co-FCM | MVKSC | MinimaxFCM | MultiNMF | MV-Co-VH |
|---|---|---|---|---|---|---|---|---|
| NMI | 0.6108 (0.0092) | 0.6147 (0.0286) | 0.6328 (0.0107) | 0.6440 (0.0067) | 0.5416 (1.17e-16) | 0.6306 (0.0204) | 0.6524 (0.0283) | **0.6673 (0.0142)** |
| RI | 0.8783 (9.96e-06) | 0.8603 (0.0150) | 0.8860 (0.0042) | 0.8835 (1.13e-05) | 0.8422 (1.17e-16) | 0.8693 (0.0098) | 0.8829 (0.0084) | **0.8902 (0.0047)** |
| Precision | 0.5726 (3.98e-05) | 0.5106 (0.0426) | 0.5985 (0.0151) | 0.5899 (5.33e-05) | 0.4552 (1.17e-16) | 0.5701 (0.0189) | 0.5832 (0.0263) | **0.6092 (0.0161)** |

Table V Performance of algorithms on the Water Treatment Plant dataset (SD in bracket).

| **Index** | FCM | CombKM | Co-FKM | WV-Co-FCM | MVKSC | MinimaxFCM | MultiNMF | MV-Co-VH |
|---|---|---|---|---|---|---|---|---|
| NMI | 0.1928 (0.0058) | 0.1903 (0.0206) | 0.2174 (0.0056) | 0.1947 (0.0121) | 0.1163 (0.0027) | 0.2048 (0.0084) | 0.1617 (0.0170) | **0.2246 (0.0037)** |
| RI | 0.7109 (0.0031) | 0.7011 (0.0098) | 0.7139 (0.0020) | 0.7112 (0.0014) | 0.6985 (1.93e-05) | 0.7137 (0.0017) | 0.7026 (0.0050) | **0.7146 (0.0054)** |
| Precision | 0.4104 (0.0118) | 0.4045 (0.0169) | 0.4266 (0.0072) | 0.4107 (0.0075) | 0.3302 (0.0035) | 0.4192 (0.0194) | 0.3726 (0.0236) | **0.4307 (0.0247)** |

Table VI Performance of algorithms on the Dermatology dataset (SD in bracket).

| **Index** | FCM | CombKM | Co-FKM | WV-Co-FCM | MVKSC | MinimaxFCM | MultiNMF | MV-Co-VH |
|---|---|---|---|---|---|---|---|---|
| NMI | 0.8688 (0.0215) | 0.8337 (0.0897) | 0.8807 (0.0036) | 0.8779 (0.0030) | 0.7121 (0.0278) | 0.8803 (0.0209) | 0.8469 (0.0039) | **0.9028 (0.0317)** |
| RI | 0.9140 (0.0099) | 0.8981 (0.0297) | 0.9310 (0.0222) | 0.9249 (0.0117) | 0.8907 (0.0014) | 0.9244 (0.0275) | 0.9069 (0.0073) | **0.9503 (0.0274)** |
| Precision | 0.8167 (1.17e-16) | 0.7427 (0.1163) | 0.8161 (0.0591) | 0.8234 (0.0164) | 0.7399 (0.0131) | 0.8082 (0.0676) | 0.8032 (0.0221) | **0.8666 (0.0747)** |

Table VII Performance of algorithms on the Forest Type dataset (SD in bracket).

| **Index** | FCM | CombKM | Co-FKM | WV-Co-FCM | MVKSC | MinimaxFCM | MultiNMF | MV-Co-VH |
|---|---|---|---|---|---|---|---|---|
| NMI | 0.5452 (0) | 0.5398 (0.0040) | 0.5618 (1.17e-16) | 0.5638 (5.05e-04) | 0.4230 (0.0092) | 0.5437 (1.04e-16) | 0.5669 (0.0294) | **0.5799 (0.0037)** |
| RI | 0.7936 (0) | 0.7928 (0.0025) | 0.8077 (0) | 0.8113 (3.75e-04) | 0.7446 (0.0079) | 0.7920 (1.17e-16) | 0.8098 (0.0204) | **0.8187 (0.0031)** |
| Precision | 0.6355 (1.17e-16) | 0.6303 (0.0041) | 0.6661 (1.17e-16) | 0.6693 (5.83e-04) | 0.5425 (0.0119) | 0.6355 (1.17e-16) | 0.6642 (0.0372) | **0.6812 (0.0063)** |

Table VIII Performance of algorithms on the Caltech dataset (SD in bracket).

| **Index** | FCM | CombKM | Co-FKM | WV-Co-FCM | MVKSC | MinimaxFCM | MultiNMF | MV-Co-VH |
|---|---|---|---|---|---|---|---|---|
| NMI | 0.5533 (9.79e-17) | 0.5607 (6.35e-04) | 0.5667 (7.40e-17) | 0.7322 (0.0109) | 0.2688 (5.85e-17) | 0.6249 (0) | 0.8803 (0.0012) | **0.9037 (8.27e-17)** |
| RI | 0.7997 (1.17e-16) | 0.8043 (5.00e-04) | 0.8041 (1.17e-16) | 0.9146 (0.0040) | 0.6442 (1.17e-16) | 0.8411 (1.17e-16) | 0.9629 (2.95e-04) | **0.9727 (5.96e-04)** |
| Precision | 0.7130 (1.17e-16) | 0.7194 (7.52e-04) | 0.7187 (0) | 0.8864 (0.0048) | 0.4981 (5.85e-17) | 0.7713 (1.17e-16) | 0.9553 (5.30e-04) | **0.9657 (6.21e-04)** |
8

Table IX Performance of algorithms on the Corel dataset (SD in bracket).

| **Index** | FCM | CombKM | Co-FKM | WV-Co-FCM | MVKSC | MinimaxFCM | MultiNMF | MV-Co-VH |
|---|---|---|---|---|---|---|---|---|
| NMI | 0.2763 (0.0030) | 0.2697 (0.0125) | 0.2646 (0.0038) | 0.2635 (0.0065) | 0.1925 (0.0030) | 0.2775 (0.0069) | 0.2815 (0.0088) | **0.2966 (0.0136)** |
| RI | 0.8449 (0.0010) | 0.8368 (0.0077) | 0.8473 (0.0015) | 0.8449 (0.0145) | 0.8301 (0.0047) | 0.8446 (0.0032) | 0.8446 (0.0019) | **0.8499 (0.0046)** |
| Precision | 0.2375 (0.0029) | 0.2278 (0.0143) | 0.2377 (0.0069) | 0.2292 (0.0042) | 0.1804 (0.0064) | 0.2363 (0.0089) | 0.2403 (0.0061) | **0.2593 (0.0188)** |

Table X Performance of algorithms on the Reuters dataset (SD in bracket).

| **Index** | FCM | CombKM | Co-FKM | WV-Co-FCM | MVKSC | MinimaxFCM | MultiNMF | MV-Co-VH |
|---|---|---|---|---|---|---|---|---|
| NMI | 0.1886 (0.0082) | 0.2084 (0.0982) | 0.1897 (0.0420) | 0.2178 (0.0188) | 0.1941 (0.0172) | 0.1938 (0.0140) | 0.2665 (0.0030) | **0.3211 (0.0267)** |
| RI | 0.6792 (0.0410) | 0.5321 (0.1633) | 0.7006 (0.0423) | 0.6983 (0.0334) | 0.7450 (0.0013) | 0.7307 (0.0461) | **0.7587 (0.0017)** | 0.7430 (0.0111) |
| Precision | 0.2287 (0.0281) | 0.2254 (0.0435) | 0.2542 (0.0397) | 0.2475 (0.0201) | 0.2409 (0.0038) | 0.2542 (0.0325) | **0.3058 (0.0030)** | 0.2964 (0.0219) |

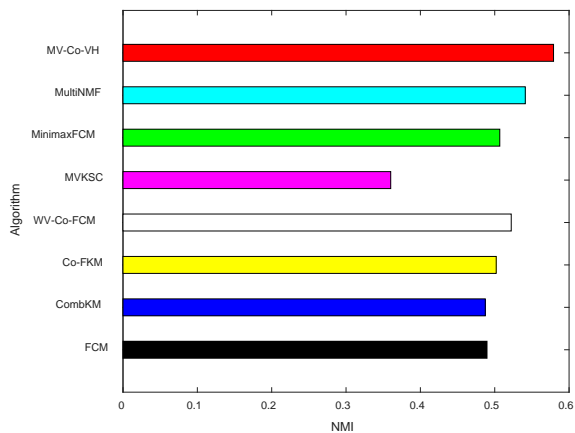

Fig. 4 Mean NMI of the algorithms on all datasets.

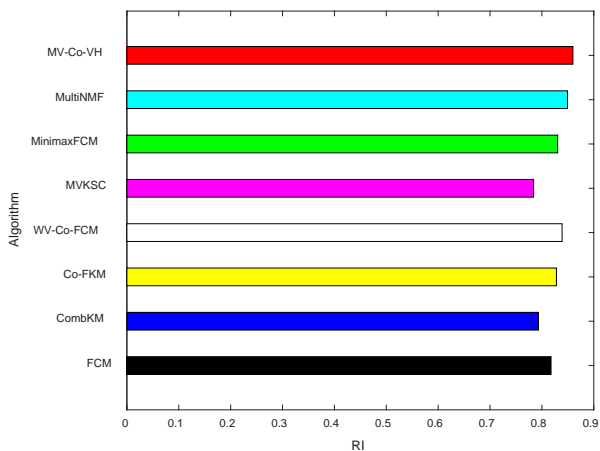

Fig. 5 Mean RI of the algorithms on all datasets.

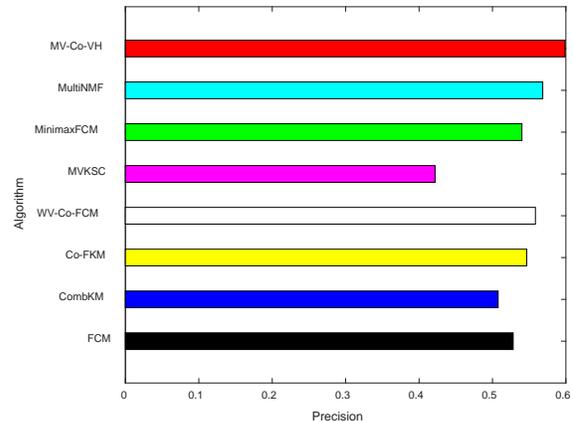

Fig. 6 Mean Precision of the algorithms on all datasets.

*C. Parameter and Convergence Analysis*

In this section, the parameters of the proposed MV-Co-VH and the convergence of the algorithm is analyzed.

*1) Parameter analysis*

The proposed MV-Co-VH algorithm is investigated to study the influence of hidden view on the clustering performance of multi-view data. The investigation is conducted by analyzing the effect of the collaborative learning parameter $\beta$ which can be adjusted to control the proportion of the visible and hidden views exploited in the clustering process. When $\beta = 0$, MV-Co-VH only uses the information of the visible views for clustering; when $\beta = 1$, MV-Co-VH only uses the information of hidden view; when $0 < \beta < 1$, MV-Co-VH utilizes both visible and hidden information to different extent. To investigate the effect of $\beta$ on the clustering performance, we conducted experiments by fixing all variables and setting $\beta$ to 11 different values respectively, i.e., $0, 0.1, 0.2, ..., 1$. It is found that the performance trends in terms of the three performance indices NMI, RI, and Precision are similar on all datasets. Hence, we present the analysis only in terms of NMI and on three of the datasets, i.e., Multiple Features, Forest Type and Caltech as an example.

Fig. 7 shows the variation of NMI with $\beta$ on the three datasets. It can be seen that the NMI values obtained by using only the visible views are not optimal on the three datasets. At the same time, when the hidden view information is introduced, the clustering performance of the proposed MV-Co-VH algorithm become better than that using the visible views information only. The clustering performance can be significantly improved with an appropriate $\beta$. This indicates that the cooperation of visible and hidden information can enhance clustering performance effectively with an optimal $\beta$.

However, how to determine the optimal value of $\beta$ is still an open problem to be further studied. In this paper, we used the grid search strategy to optimize this parameter.

*2) Convergence Analysis*

In the proposed MV-Co-VH algorithm, we use an alternate iteration method to minimize the objective function. When the change in the value of objective function is smaller than a given threshold, the iteration stops. Fig. 8 shows the variation of the algorithm of the objective function with the number of iterations for some datasets. It can be seen that the algorithms reach convergence after a certain number of iterations for all cases. In addition, the convergence of the algorithm is guaranteed based on the Zangwill convergence theorem [33, 34].

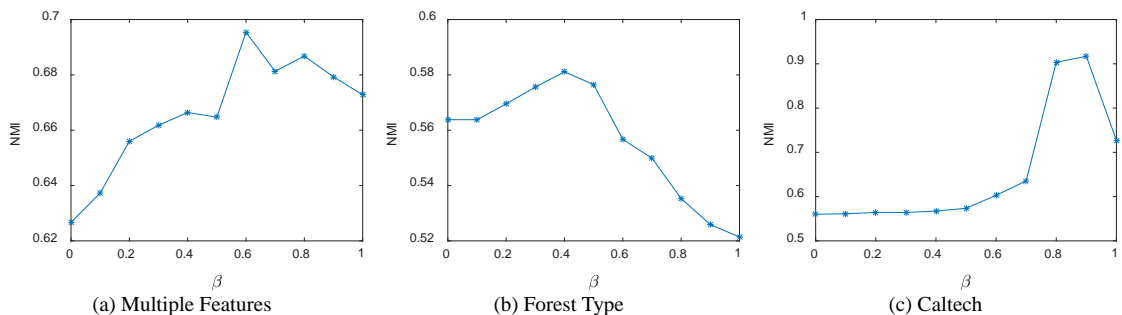

Fig. 7 Variation of the performance of MV-Co-VH with $\beta$.

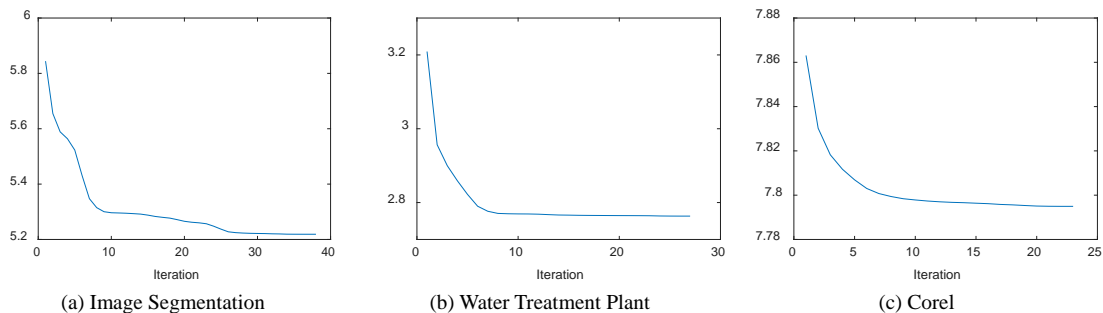

Fig. 8 Convergence of MV-Co-VH.

### D. Effect of the Hidden Information

In order to verify whether the addition of hidden information can improve the performance of the multi-view clustering algorithm, this section compares the clustering results of the proposed algorithm MV-Co-VH with and without the hidden information. The specific results based on NMI are shown in Table XI. By observing the clustering results of Table XI, it can be seen that the introduction of the hidden information helps to improve the clustering performance on most datasets.

Table XI Performance of MV-Co-VH with and without the hidden information.

| Dataset | MV-Co-VH (without the hidden information) | MV-Co-VH (with the hidden information) |
|---|---|---|
| Multiple Features | 0.6943 (0.0130) | **0.7369 (0.0091)** |
| Image Segmentation | 0.6253 (0.0258) | **0.6673 (0.0142)** |
| Water Treatment Plant | 0.1606 (0.0850) | **0.2246 (0.0037)** |
| Dermatology | 0.8900 (0.0464) | **0.9028 (0.0317)** |
| Forest type | 0.5637 (0.0069) | **0.5799 (0.0037)** |
| Caltech | 0.7284 (0.1179) | **0.9037 (8.27e-17)** |
| Corel | 0.2741 (0.0120) | **0.2966 (0.0136)** |
| Reuters | 0.2529 (0.0821) | **0.3211 (0.0267)** |

11*E. Statistical Analysis*

In order to determine whether the performance improvement of the proposed MV-Co-VH algorithm observed is significant or not, the Friedman test [35, 36], a non-parametric statistical analysis method, is used to analyze the experimental results of the eight algorithms for all the datasets. We set the significance level $\alpha$ to 0.05. If the p-value is less than $\alpha$, the null hypothesis that all algorithms have the same performance is rejected, i.e., the performance of the algorithms is significantly different. In that case, the Holm post-hoc test is then used to further evaluate the performance difference between the best algorithm and the other algorithms. Since the performance trends in terms of the three performance indices NMI, RI, and Precision are similar, the statistical analysis is presented only with NMI as an example. The results are shown in Table XII and Table XIII.

Table XII shows the Friedman test results based on NMI. It is evident that there are significant performance differences among the eight algorithms. Furthermore, it can be seen from the ranking that (the lower the better) the proposed MV-Co-VH algorithm ranked first and thus the best algorithm. To further determine whether the performance improvement observed in MV-Co-VH (the best algorithm) is significantly better, the Holm post-hoc tests were conducted between MV-Co-VH and each of the other seven algorithms respectively. The results shown in Table XIII indicate that there is significant difference in performance between MV-Co-VH and seven of the algorithms, i.e., FCM, CombKM, WV-Co-FCM, Co-FKM, MVKSC, MinimaxFCM and MultiNMF.

Table XII Friedman Test analysis based on NMI.

| Algorithm | Ranking |
|---|---|
| FCM | 5.875 |
| CombKM | 5.75 |
| Co-FKM | 4 |
| WV-Co-FCM | 3.875 |
| MVKSC | 7.625 |
| MinimaxFCM | 4.125 |
| MultiNMF | 3.75 |
| MV-Co-VH | 1 |

Table XIII Holm post-hoc test analysis based on NMI.

| *i* | Algorithm | $z=(R_0-R_i)/SE$ | *p*-value | Holm=$\alpha/i$ | Hypothesis |
|---|---|---|---|---|---|
| 7 | MVKSC | 5.40929 | 0 | 0.007143 | Rejected |
| 6 | FCM | 3.980421 | 0.000069 | 0.008333 | Rejected |
| 5 | CombKM | 3.878359 | 0.000105 | 0.01 | Rejected |
| 4 | MinimaxFCM | 2.551552 | 0.010724 | 0.0125 | Rejected |
| 3 | Co-FKM | 2.44949 | 0.014306 | 0.016667 | Rejected |
| 2 | WV-Co-FCM | 2.347428 | 0.018904 | 0.025 | Rejected |
| 1 | MultiNMF | 2.245366 | 0.024745 | 0.05 | Rejected |

V. CONCLUSION AND FUTURE WORK

In this paper, we propose the multi-view clustering algorithm MV-Co-VH with the cooperation of visible and hidden views based on the classical K-means algorithm framework. We first develop a method to extract the shared hidden view data with NMF technique. Then, by introducing hidden view into the clustering process, collaborative learning between the visible and hidden views of multi-view data is implemented. Thus, the proposed algorithm not only utilizes the otherness information between different views, but also the consistency information among them. The experimental results on UCI multi-view datasets and real-world image multi-view datasets show that the proposed algorithm has better clustering performance than that of the existing clustering algorithms. Further research will be conducted on two aspects: how to determine the optimal collaborative learning parameter, and how to optimize the hyper-parameters more efficiently.

REFERENCES

[1] J. A. Hartigan, "A K-Means Clustering Algorithm," *Appl Stat*, vol. 28, no. 1, pp. 100-108, 1979.
[2] L. Jing, M. K. Ng, and J. Z. Huang, "An Entropy Weighting k-Means Algorithm for Subspace Clustering of High-Dimensional Sparse Data," *IEEE Transactions on Knowledge & Data Engineering*, vol. 19, no. 8, pp. 1026-1041, 2007.
[3] S. Yu *et al*, "Optimized data fusion for kernel k-means clustering," *IEEE Transactions on Pattern Analysis and Machine Intelligence*, vol. 34, no. 5, pp. 1031-1039, 2012.
[4] J. C. Bezdek, R. Ehrlich, and W. Full, "FCM: The fuzzy c-means clustering algorithm," *Computers & Geosciences*, vol. 10, no. 2, pp. 191-203, 1984.
[5] L. Zhu, F. L. Chung, and S. Wang, "Generalized Fuzzy C-Means Clustering Algorithm With Improved Fuzzy Partitions," *IEEE Transactions on Systems Man & Cybernetics Part B Cybernetics*, vol. 39, no. 3, pp. 578-591, 2009.
[6] L. O. Hall and D. B. Goldgof, "Convergence of the Single-Pass and Online Fuzzy C-Means Algorithms," *IEEE Transactions on Fuzzy Systems*, vol. 19, no. 4, pp. 792-794, 2011.
[7] M. Ester, H. P. Kriegel, J. Sander, and X. Xu, "A density-based algorithm for discovering clusters in large spatial databases with noise," in *Proceedings of the Second International Conference on Knowledge Discovery and Data Mining*, 1996, pp. 226–231.